\pdfoutput=1

\documentclass[11pt]{article}

\usepackage{acl}
\usepackage{times}
\usepackage{latexsym}

\usepackage[T1]{fontenc}

\usepackage[utf8]{inputenc}

\usepackage{microtype}

\usepackage{inconsolata}

\definecolor{myblue}{HTML}{D0E6FD}
\usepackage{graphicx}
\usepackage{booktabs}
\usepackage{algorithm}
\usepackage{algpseudocode}
\usepackage{multicol}
\usepackage{tcolorbox}
\usepackage{array}
\usepackage{geometry}
\usepackage{multirow}
\usepackage{multicol}

\title{KUNPENG: An Embodied Large Model for Intelligent Maritime}

\author{%
  Naiyao Wang$^{1}$, \quad
  Tongbang Jiang$^{1}$, \quad
  Ye Wang$^{1}$, \quad
  Shaoyang Qiu$^{1}$, \\ 
  \textbf{Bo Zhang}$^{1}$, \quad
  \textbf{Xinqiang Xie}$^{1}$, \quad
  \textbf{Munan Li}$^{1}$, \quad
  \textbf{Chunliu Wang}$^{1}$, \quad
  \textbf{Yiyang Wang}$^{1}$, \quad \\
  \textbf{Hongxiang Ren}$^{2}$, \quad
  \textbf{Ruili Wang}$^{1}$, \quad
  \textbf{Hongjun Shan}$^{3}$, \quad
  \textbf{Hongbo Liu}$^{1}$\thanks{Corresponding author: \texttt{lhb@dlmu.edu.cn} }\\
  $^{1}$College of Artificial Intelligence, Dalian Maritime University \\
  $^{2}$Navigation College, Dalian Maritime University \\
  $^{3}$Law School, Dalian Maritime University\\
    Project Website: \url{https://github.com/ACoTAI/KUNPENG}
}

\begin{document}
\maketitle
\begin{abstract}
Intelligent maritime, as an essential component of smart ocean construction, deeply integrates advanced artificial intelligence technology and data analysis methods, which covers multiple aspects such as smart vessels, route optimization, safe navigation, aiming to enhance the efficiency of ocean resource utilization and the intelligence of transportation networks. However, the complex and dynamic maritime environment, along with diverse and heterogeneous large-scale data sources, present challenges for real-time decision-making in intelligent maritime. In this paper, We propose KUNPENG, the first-ever embodied large model for intelligent maritime in the smart ocean construction, which consists of six systems. The model perceives multi-source heterogeneous data for the cognition of environmental interaction and make autonomous decision strategies, which are used for intelligent vessels to perform navigation behaviors under safety and emergency guarantees and continuously optimize power to achieve embodied intelligence in maritime. In comprehensive maritime task evaluations, KUNPENG has demonstrated excellent performance.
\end{abstract}
\section{Introduction}
Intelligent maritime, a cutting-edge integration of artificial intelligence (AI) with maritime operations, has revolutionized the traditional maritime industry~\citep{Qu2023Multi,Zenia2023Reer}. As the global maritime industry continues to expand, the adoption of intelligent maritime systems can optimize route planning, monitor vessel conditions in real-time, and facilitate autonomous navigation~\citep{Sui2022Node,Zhang2021Collision}. This evolution is driven by the need to reduce operational costs, comply with stricter environmental regulations, and improve the overall safety of maritime activities~\citep{Zhou2023Exploring,Peng2023Safety}.

\begin{figure}[!htpb]
	\begin{center}
	    \includegraphics[width=0.49\textwidth]{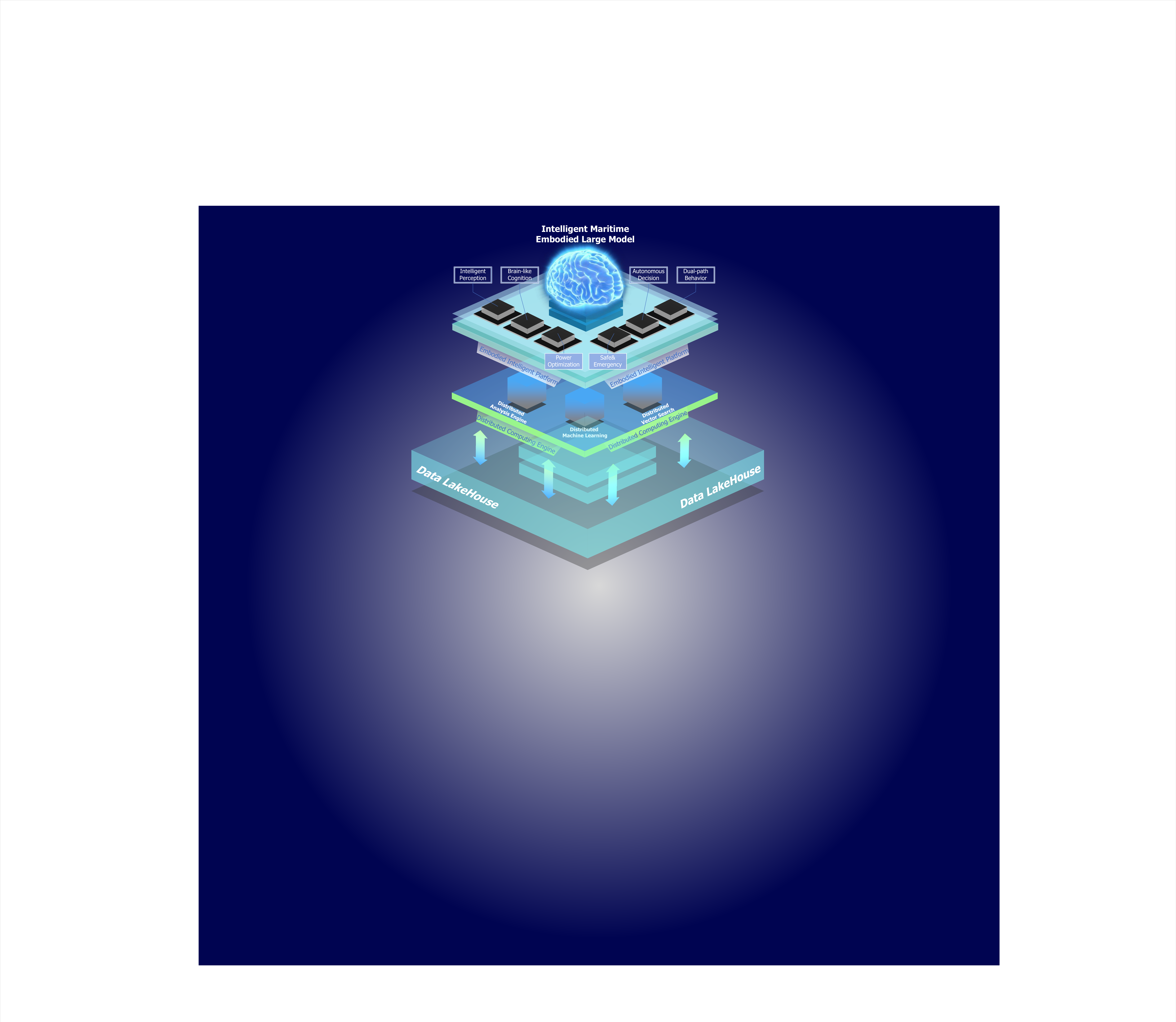}
		\caption{Overall architecture of the KUNPENG embodied large model.}
		\label{fig:model}
	\end{center}
\end{figure}

In recent years, the intelligent maritime sector has witnessed significant technological advancements, including the development of autonomous ships, enhanced AI-driven route optimization algorithms, and sophisticated predictive maintenance systems~\citep{Zhang2022Vessel}. These innovations are complemented by the integration of Internet of Things (IoT) devices and advanced communication networks, which together facilitate real-time data collection and analysis~\citep{Ashraf2022Survey,Liu2022Deep,Liu2022Intelligent}.

However, despite the promising advancements, the field of intelligent maritime faces several critical challenges. One of the primary issues is the integration and processing of heterogeneous data from various sources, such as sensors, satellite communications, and weather forecasts~\citep{Liu2022Moving,Liu2022Intrusion}. This complexity often leads to fragmented and suboptimal decision-making processes. Additionally, the dynamic and unpredictable nature of maritime environments poses significant hurdles for real-time data analysis and adaptive responses~\citep{Lin2022Maritime,Zhang2021Hkpm,Clunie2021Development}.

In this work, an embodied large model named KUNPENG is proposed for intelligent maritime, emphasizing the interaction between computational models and physical environment to enhance the stability and adaptability of intelligent maritime systems. By integrating multimodal data with distributed computing and providing a more holistic understanding of the maritime environment with embodied platform, KUNPENG endows intelligent vessels with AI brains, ensuring they can autonomously and compliantly execute complex marine tasks in dynamic environment. KUNPENG first enables intelligent vessels to perceive heterogeneous data in real-time, facilitating the cognition of embodied environmental interactions and autonomous navigation decision-making. Subsequently, it enhances maritime safety and emergency response capabilities for compliant navigation, ultimately optimizing navigation behaviors and power configurations to achieve the embodied intelligent maritime for smart ocean construction, which is shown in Figure~\ref{fig:model}.

To sum up, the main contributions of this work are as follows:
\begin{itemize}
\item \noindent We propose the first-ever embodied large model named KUNPENG for intelligent maritime, designed to perceive multi-source heterogeneous data for the cognition of environmental interactions and to make autonomous decisions. The model enables intelligent vessels to perform navigation behaviors under safety and emergency guarantees while continuously optimizing power usage, ultimately achieving embodied intelligence in maritime and enhancing both operational efficiency and safety.
\item \noindent We propose an embodied intelligent loop of "Perception-Cognition-Decision-Behavior" based on a brain-like dual-pathway model, utilizing Theory of Mind and Game Theory to achieve cross-domain group collaborative intelligent maritime in complex scenarios. The loop can perceive environmental dynamics more accurately like the human brain, understand and coordinate behaviors among multiple vessels, which ensures fast response to actions while also forming slow-paced deep thinking experiences, achieving real-time decision adjustments and long-term behavioral optimization.
\item \noindent Vessel ego-motion imitation learning is proposed for intelligent maritime navigation. By enhancing the data quality of marine hydro-meteorological forecasts and ego-motion navigation data, vessels perceive the dynamic navigation environment embodiedly  and combines virtual simulation with physical embodiment to mimic expert navigation decisions and behaviors, thereby constructing an integrated embodied meteorological navigation scheme.
\item \noindent We propose safety navigation rules and standards under intelligent maritime by constructing a credit graph in the maritime domain. The graph integrates structured and unstructured credit data, deeply exploring semantic relationships among credit entities in maritime. Leveraging maritime law, we achieve compliant and safe navigation for intelligent vessels. Furthermore, feature fusion is accomplished through multimodal data representation learning, enabling intelligent maritime surveillance reports and alerting of hazardous scenarios.

\end{itemize}

\begin{figure*}[!t]
	\begin{center}
	    \includegraphics[width=1\textwidth]{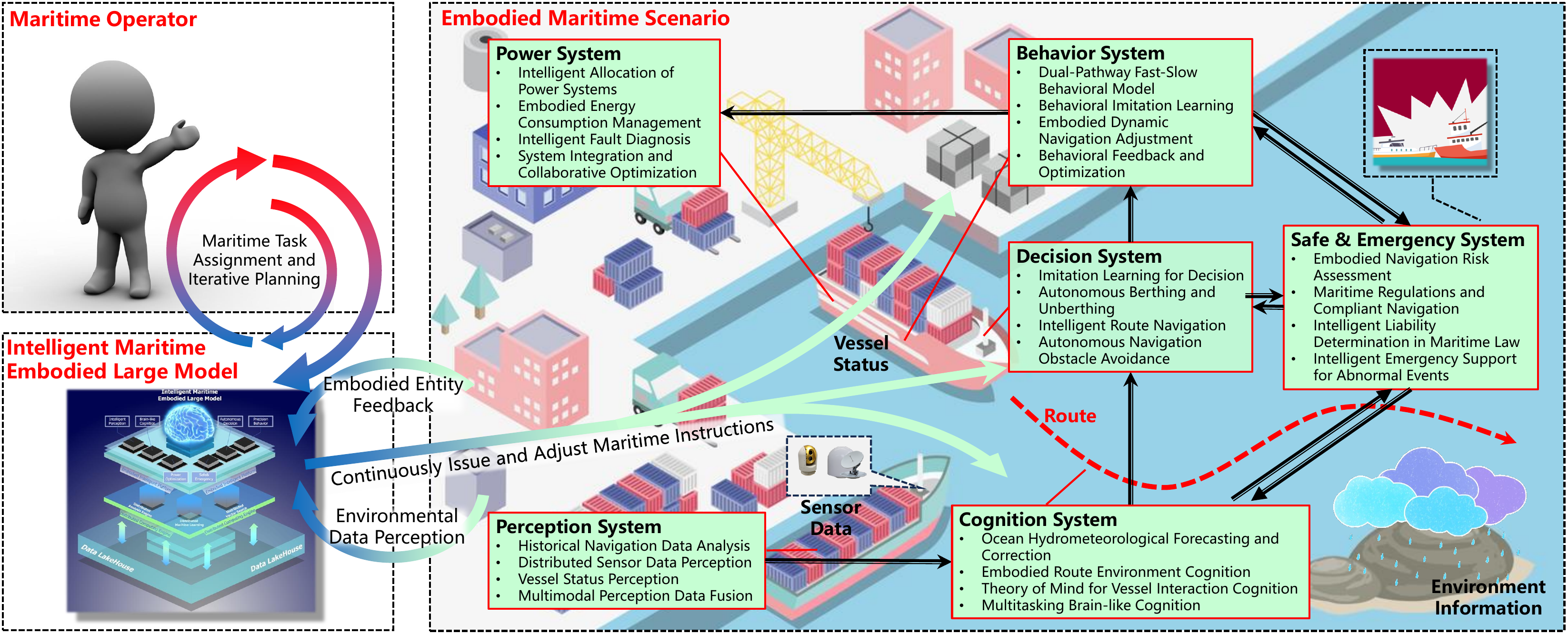}
		\caption{Intelligent maritime of the KUNPENG embodied large model, including Perception System, Cognition System, Decision System, Behavior System, Power System, and Safe \& Emergency System.}
		\label{fig:architecture}
	\end{center}
\end{figure*}

\section{Related Work}
\paragraph{General Large Models}

General large models have seen widespread development, encompassing areas such as language, vision, generation, and recommendation, providing efficient solutions for various complex tasks~\citep{Chang2023Survey}. Majumdar et al. present a modern formulation of Embodied Question Answering (EQA) as the task of understanding an environment well enough to answer questions about it in natural language~\citep{Majumdar2024Openeqa}. Yang et al. train a vision-language model (VLM) agent living in a visual world using an LLM agent excelling in a parallel text world~\citep{Yang2024Embodied}. The development of large models has also gradually expanded into the ocean domain.~\citep{Bi2023Oceangpt} introduce OceanGPT, the frst-ever large language model in the ocean domain, which is expert in various ocean science tasks.~\citep{Yang2023Oceanchat} propose OceanChat, a system that leverages a closed-loop LLM-guided task and motion planning framework to tackle Autonomous Underwater Vehicle (AUV) missions in the wild.~\citep{Yang2024Oceanplan} develop a hierarchical LLM-task-motion planning and replanning framework to efficiently ground an abstracted human command into tangible AUV control.

\paragraph{Embodied Large Models}
To achieve environmental interaction and cognition, thereby enhancing autonomous decision-making and behavioral capabilities in complex dynamic environments, embodied intelligence injects new vitality into general large models.~\citep{Zheng2024Towards} propose the first generalist model for embodied navigation NaviLLM. It adapts LLMs to embodied navigation by introducing schema-based instruction.~\citep{Nottingham2023Embodied} focus on the task of creating embodied RL agents that can exploit large-scale external knowledge sources presented in the form of pretrained large language models.~\citep{Song2023Llm} propose a novel method, LLM-Planner, that harnesses the power of large language models to do few-shot planning for embodied agents.~\citep{Dorbala2023Can} present language-guided exploration (LGX), an embodied agent navigates to an uniquely described target object in a previously unseen environment.~\citep{Yang2023Harnessing} consider that Truly embodying AI using LLMs is a stepping stone to achieving artificial general intelligence.~\citep{Wang2024Describe} investigate the challenge of task planning for multi-task embodied agents and propose an interactive planning approach based on Large Language Models (LLMs). In conclusion, with the extension of large models into various fields, the demand for environmental interaction and real-time decision-making makes embodied intelligence in maritime large models an inevitable trend.
\section{KUNPENG}
To achieve more efficient environmental interaction and real-time autonomous decision-making, thereby enhancing maritime safety and operational efficiency, we propose an intelligent maritime embodied large model (KUNPENG). The overall architecture of the KUNPENG model (see Figure~\ref{fig:architecture}) has six parts: The perception system senses and integrates heterogeneous data on vessel status, environmental conditions, and historical navigation, which the cognition system uses to predict and correct oceanic meteorological states, understand the interactions between the environment and the vessel, and form brain-like cognition for multi-tasks in maritime. This provides a foundation for the decision-making system to make Theory of Mind (ToM) decisions, enabling intelligent vessels to obtain optimal strategies for autonomous berthing and unberthing, navigation, and obstacle avoidance. These strategies will serve as the basis for the dual-pathway fast-slow behavioral model in the behavior system, achieving embodied dynamic maritime behavior through feedback and optimization via imitation learning. The power system intelligently manages energy consumption and diagnoses faults during maritime behavior, ensuring collaborative optimization across systems. The safety and emergency system provides comprehensive risk assessments throughout the maritime process, ensuring safe and compliant navigation based on maritime regulations, and in case of anomalies, it determines responsibility based on maritime law to support intelligent emergency handling.

\begin{figure*}[!t]
	\begin{center}
	    \includegraphics[width=1\textwidth]{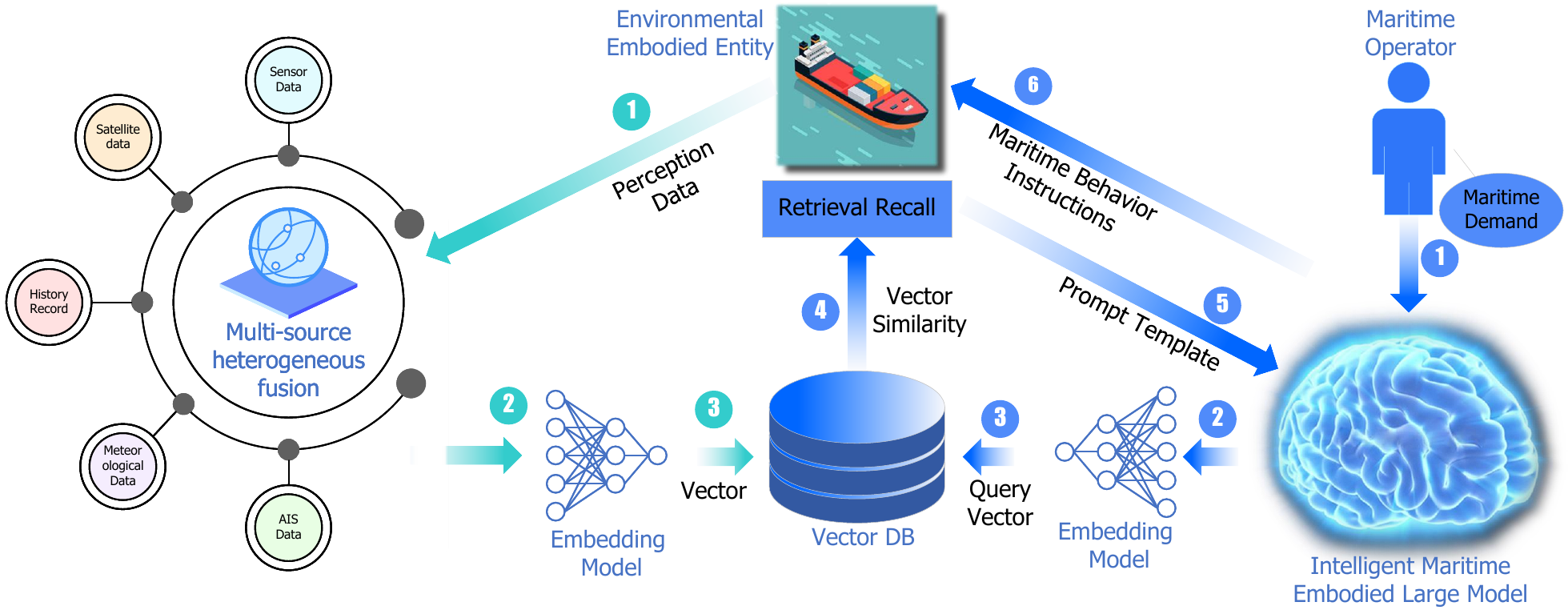}
		\caption{Intelligent maritime of the KUNPENG embodied large model.}
		\label{fig:train}
	\end{center}
\end{figure*}
\subsection{Workflow of the KUNPENG Model}
The KUNPENG model integrates embodied artificial intelligence and multimodal data fusion technologies, aimed at handling maritime demand queries and generating autonomous navigation instructions for safe and compliant navigation of intelligent vessels. The overall workflow of the KUNPENG model is briefly presented as follows, which is shown in Figure~\ref{fig:train}:
\begin{itemize}
\item \textbf{Step-1 Demand Input:} Maritime operators input queries describing maritime tasks, including origin, destination, cargo type, and transport time.
\item \textbf{Step-2 Data Encoding:} Preprocessing of maritime demand queries includes tokenization, part-of-speech tagging, and semantic understanding to accurately capture the intent of the maritime operators. At the same time, integrating sensor data of environmental status, satellite data, historical navigation data, ocean meteorological data, and Automatic Identification System (AIS) data of vessels status into vectors, capturing their semantic and feature information.
\item \textbf{Step-3 Retrieval:} The model retrieves relevant data from an established maritime knowledge base, including route planning, nautical charts, weather forecasts, and channel conditions. This information is crucial for autonomous vessel navigation and task execution.
\item \textbf{Step-4 Candidate Information Fusion:}: The retrieved related vectors are fused with the query vectors. This process may involve the integration of multimodal information, combining features of text and image data to generate a comprehensive representation.
\item \textbf{Step-5 Generation:} Leveraging the fused vectors, specific maritime instructions are generated. These instructions encompass route planning, speed adjustments, navigational avoidance, target arrival times, etc., ensuring safe and efficient autonomous execution of maritime tasks by intelligent vessels.
\item \textbf{Step-6 Instruction Output:} Generated maritime instructions are output in machine-readable formats for consumption by autonomous systems onboard intelligent vessels. These instructions may directly input into navigation or control systems to guide vessels according to designated maritime plans and strategies.
\end{itemize}

\subsection{Main Systems}
\paragraph{Intelligent Perception}
The intelligent perception system of the embodied intelligent maritime model integrates real-time data from vessel status, environmental conditions, and historical navigation to provide a comprehensive understanding of the maritime environment~\citep{Li2023Sense,Wang2021Input,Wang2023Pre}. First, through Historical Navigation Data Analysis, the system extracts valuable information and patterns from past navigation records to predict and optimize current navigation routes. Second, Distributed Sensor Data Perception involves real-time collection of environmental data through various sensors deployed on the vessel, such as radar, sonar, and meteorological sensors, ensuring comprehensive awareness of the surrounding environment~\citep{Wang2024Soft,Sun2017Substructural}. Third, Vessel Status Perception continuously monitors the vessel's operational status, including key parameters like engine performance, fuel levels, speed, and heading, to ensure safe and normal operation. Lastly, Multimodal Perception Data Fusion integrates and analyzes data from various sources and types, forming a comprehensive situational awareness map that provides more accurate and thorough perception information, supporting efficient decision-making and operations in the intelligent maritime system.
\paragraph{Brain-like Cognition}
The brain-like cognition system refers to the simulation of human brain thinking and learning processes through complex computational models to achieve advanced understanding and decision-making in environments and tasks. Ocean Hydrometeorological Forecasting and Correction allows the system to accurately predict oceanic conditions and dynamically adjust based on real-time data. Building on this, Embodied Route Environment Cognition integrates multi-source perception data to deeply understand and recognize changes and characteristics in the route environment~\citep{Wang2023Pixels}. This understanding is further enhanced by Theory of Mind for Vessel Interaction Cognition, which endows the system with the ability to infer the intentions and behaviors of other vessels, optimizing interactions and coordination between them~\citep{Liu2023Semi}. Finally, Multitasking Brain-like Cognition allows the system to handle multiple tasks simultaneously, making intelligent decisions and operations in complex environments~\citep{Guo2023Domain}. Through these interconnected functions, the brain-like cognition system enhances the maritime system's ability to respond and operate efficiently in complex and dynamic environments.
\paragraph{Autonomous Decision}
Autonomous decision-making refers to the system's capability to independently analyze and comprehend complex information based on environmental perception and prediction, enabling effective decision-making and action execution. Imitation Learning for Decision allows the system to learn decision-making strategies from expert behavior, ensuring effective operations and responses~\cite{Wang2022Fine}. This understanding directly informs Autonomous Berthing and Unberthing operations, where the system autonomously navigates through various berth conditions, adapting dynamically to ensure safe and efficient docking maneuvers. Simultaneously, Intelligent Route Navigation leverages real-time data and advanced algorithms to select optimal routes, considering dynamic environmental factors and vessel interactions identified through Theory of Mind~\citep{Wang2022Seem}. This comprehensive approach not only enhances operational efficiency but also ensures navigational safety by integrating Autonomous Navigation Obstacle Avoidance, which continuously monitors and avoids obstacles in real-time, safeguarding smooth and uninterrupted vessel movement~\citep{Wang2024Duel}. Together, these interconnected components form a robust autonomous decision-making system that enhances overall maritime operations by seamlessly integrating perception, prediction, and adaptive navigation capabilities.
\paragraph{Dual-path Behavior}
The Dual-Pathway Fast-Slow Behavioral Model allows the system to balance between rapid responses and deliberate decision-making processes, adapting to varying operational needs efficiently. Complementing this, Behavioral Imitation Learning refines behavioral patterns and best practices through experiential learning, enhancing the system's adaptability and performance in diverse maritime scenarios. Furthermore, Embodied Dynamic Navigation Adjustment continuously adjusts navigation routes based on real-time data inputs, ensuring responsive navigation in dynamic and unpredictable environments~\citep{Qiu20233d}. This capability is reinforced by Behavioral Feedback and Optimization, which utilizes feedback mechanisms to continually optimize behavioral strategies and decision-making processes, enhancing operational efficiency and safety.
\paragraph{Power Optimization}
Power Optimization in KUNPENG is a comprehensive system aimed at optimizing the utilization of electrical resources and enhancing system efficiency through multifaceted strategies and technologies. Firstly, an Intelligent Allocation of Power Systems allocates electrical resources by real-time monitoring and analyzing the power demands of various ship components, ensuring optimal performance and efficiency across different operational states. Secondly, an Embodied Energy Consumption Management system integrates multi-source data and intelligent algorithms to dynamically optimize energy consumption, effectively controlling vessel energy expenditure to extend sailing time and enhance navigation stability. Additionally, the system employs Intelligent Fault Diagnosis technology to monitor the operational status and health of the power system in real-time, promptly identifying and locating potential faults, and implementing preventive or corrective measures to ensure reliable power supply under any circumstances. Finally, System Integration and Collaborative Optimization ensures harmonious operation among subsystems, enhancing overall efficiency and performance stability by optimizing system structure and processes, thereby improving the safety and cost-effectiveness of maritime operations.
\paragraph{Safe \& Emergency}
Safety and Emergency Systems refer to the subsystems within the intelligent maritime embodied large model responsible for assessing and responding to navigation risks, ensuring compliance with maritime regulations, and providing intelligent emergency support. Firstly, through the Embodied Navigation Risk Assessment system, the model can real-time analyze and predict various potential risks during navigation, enabling proactive measures to ensure sailing safety. Secondly, the system adheres to Maritime Regulations and Compliant Navigation, conducting compliance monitoring and management of vessel navigation behavior to ensure maritime activities operate within legal and safety frameworks. Additionally, leveraging Intelligent Liability Determination in Maritime Law, the system intelligently handles abnormal events, swiftly and accurately determining responsibilities and providing corresponding emergency support, enabling vessels to respond appropriately in emergencies. Finally, through Intelligent Emergency Support for Abnormal Events, the system effectively addresses various unexpected situations, including but not limited to weather changes, technical failures, or human errors, safeguarding the safety of vessels and their crews, ensuring smooth and continuous maritime operations.


\section{Conclusions and Future Work}
In this paper, we have proposed KUNPENG, an embodied large model for intelligent maritime that tackles the problem of the real-time decision-making in complex and dynamic maritime environment, along with diverse and heterogeneous large-scale data sources. Within the KUNPENG model, we introduce the embodied intelligent based on the brain-like dual-pathway model for perceiving environmental dynamics like the human brain, understand and coordinate behaviors among multiple vessels, and the fast path ensures real-time dynamic decision-making, and the slow path enables the accumulation of behavioral experience. Vessel ego-motion imitation learning is then proposed for constructing an integrated embodied meteorological navigation scheme. Finally, we propose the safety navigation rules and standards under intelligent maritime vessels for achieving compliant and safe navigation.

The KUNPENG model is developed as the first-ever embodied large model for intelligent maritime empowered by embodied intelligence to encompass the entire "Perception-Cognition-Decision-Behavior" process, ensuring the safe and compliant navigation of intelligent vessels in accordance with maritime law. Despite the significant potential of the embodied large model for intelligent maritime in enhancing the efficiency and safety of intelligent vessel navigation, the stability and reliability of the system under complex and extreme marine environments require further validation. Future development directions include optimizing algorithms and data processing techniques to improve the model's real-time performance and accuracy, enhancing the model's adaptability to different marine environments and meteorological conditions, and advancing the integration of virtual simulations with physical verification to build a more comprehensive and reliable intelligent maritime solution. These efforts will contribute to achieving a safer, more efficient, and intelligent global maritime network for smart ocean construction.


\section*{Acknowledgments}
We would like to extend our sincere gratitude to the Meta AI and DeepMind team for their support and contributions. The LLaMA3~\cite{Touvron2023Llama} and Perceiver~\cite{Jaegle2021Perceiver} model provided a robust foundation for our intelligent maritime embodied large model.

This work is supported
in part by the National Natural Science Foundation of China (Grant Nos. 62176036, 61772102),
in part by the Liaoning Funds (Grant No. 2023JH6/100100072),
and in part by the Fundamental Research Funds for the Central Universities (Grant No. 3132023523).

\end{document}